%% file: main-qce.tex
\RequirePackage{silence}
\documentclass[10pt,conference]{IEEEtran}
\usepackage{verbatim}
\input{preamble-common}
\input{metadata}
\input{preamble-ieee}

\begin{document}

\title{\papertitle}

\author{
\IEEEauthorblockN{Bayram Y\"{u}ksel Eker~\orcid{0009-0003-0167-5763}}
\IEEEauthorblockA{Neura Parse Ltd.\\London, United Kingdom}
\and
\IEEEauthorblockN{Suayb S.~Arslan~\orcid{0000-0003-3779-0731}}
\IEEEauthorblockA{Dept.\ of Computer Engineering\\Bo\u{g}azi\c{c}i University\\Istanbul, Turkey}
\and
\IEEEauthorblockN{\"{O}zg\"{u}r Nazl{\i}~\orcid{0009-0009-0784-0823}}
\IEEEauthorblockA{Dept.\ of Physics\\Izmir University of Economics\\Izmir, Turkey}
\linebreakand
\IEEEauthorblockN{Mustafa Serhat Demirgil~\orcid{0009-0006-4310-3546}}
\IEEEauthorblockA{Faculty of Engineering\\University of Toronto\\Toronto, Canada}
\and
\IEEEauthorblockN{Furkan Delig\"{o}z~\orcid{0009-0008-7277-8408}}
\IEEEauthorblockA{Istanbul Technical University\\Istanbul, Turkey}
}

\maketitle

\begin{abstract}
\paperabstract
\end{abstract}


\input{sections/01-introduction}

\input{sections/02-background}
\input{sections/03-multistep-grover}
\input{sections/04-biqae-calibration}
\input{sections/06-nisq-feasibility}
\input{sections/07-conclusion}

\bibliographystyle{IEEEtran}
\IEEEtriggeratref{31}
{\hbadness=3000\relax\bibliography{references}}

\end{document}

%% file: preamble-common.tex
\usepackage{amsmath}
\usepackage{amssymb}
\usepackage{mathtools}
\usepackage{bm}

\usepackage{booktabs}
\usepackage{graphicx}
\usepackage{makecell}
\usepackage{multirow}

\usepackage{tikz}
\usetikzlibrary{arrows.meta, positioning, fit, calc, decorations.pathreplacing,
                 shapes.geometric, backgrounds}
\usepackage{pgfplots}
\pgfplotsset{compat=1.18}

\usepackage{siunitx}

\usepackage{xcolor}
\definecolor{quantum-blue}{RGB}{31,119,180}
\definecolor{classical-red}{RGB}{214,39,40}
\definecolor{dwave-green}{RGB}{44,160,44}
\definecolor{nighthawk-purple}{RGB}{148,103,189}

\usepackage{microtype}   
\usepackage{pifont}      
\newcommand{\orcid}[1]{\href{https://orcid.org/#1}{\textcolor{quantum-blue}{ORCID}}}

\newcommand{\POMDP}{POMDP}

\newcommand{\MTDA}{MTDA}
\newcommand{\QBRL}{QBRL}
\newcommand{\BIQAE}{BIQAE}
\newcommand{\QANTIS}{QANTIS}

\newcommand{\OAA}{OAA}       
\newcommand{\FPAA}{FPAA}     
\newcommand{\EPLG}{EPLG}     

\newcommand{\cmark}{\ding{51}}   
\newcommand{\xmark}{\ding{55}}   

%% file: metadata.tex
\newcommand{\papertitle}{\QANTIS{}: Hardware--Calibrated Sequential \POMDP{} Belief Updates on IBM Heron}
\newcommand{\papertitleplain}{QANTIS: Hardware-Calibrated Sequential POMDP Belief Updates on IBM Heron}
\newcommand{\paperauthors}{Bayram Yuksel Eker; Suayb S. Arslan; Ozgur Nazli; Mustafa Serhat Demirgil; Furkan Deligoz}
\newcommand{\papersubject}{arXiv preprint}

\newcommand{\paperabstract}{%
Autonomous systems under partial observability act on beliefs, not raw
sensor events.  \QANTIS{} treats the quantum processor as a calibrated
belief-update service in that loop: it receives a prior and an
observation model, estimates the rare-event evidence term, and returns an
ordinary posterior to a classical planner.  This paper asks whether that
service can be reused across a sequential Tiger \POMDP{} horizon on
present IBM Heron hardware without corrupting the planner-facing
posterior.  We answer with a controlled hardware case study rather than
an end-to-end autonomy or wall-clock speedup claim.  The study compares
no amplification, guarded Grover amplification, and all-step
fixed-point amplification on the same trajectory, then checks whether
the returned posterior would change the downstream action.  All-step
\FPAA{} preserves the Tiger posterior across the reported 8-step and
12-step primary runs, and the 20-step and 32-step controls remain inside
the same operating band.  In every reported decision check, the hardware
posterior and the exact Bayes posterior select the same immediate
action.  Boundary-aware \BIQAE{} stabilizes amplitude estimation near
zero and near one, while a rare-event sweep maps the logical
sample-complexity envelope for one-in-a-million evidence.  The result is
an operating envelope for a hardware-calibrated belief-update primitive,
not a standalone hardware-advantage claim.%
}

\newcommand{\paperkeywords}{quantum computing, POMDP inference, belief updating, amplitude amplification, Bayesian amplitude estimation, NISQ hardware, IBM Heron}

%% file: preamble-ieee.tex
\usepackage{cite}
\usepackage[caption=false]{subfig}
\usepackage{hyperref}
\hypersetup{
  hidelinks,
  pdftitle={\papertitleplain},
  pdfauthor={\paperauthors},
  pdfsubject={\papersubject},
  pdfkeywords={\paperkeywords}
}
\usepackage{orcidlink}
\renewcommand{\orcid}[1]{\orcidlink{#1}}

\makeatletter
\newcommand\linebreakand{%
  \end{@IEEEauthorhalign}%
  \hfill\mbox{}\par%
  \mbox{}\hfill\begin{@IEEEauthorhalign}%
}
\makeatother

%% file: sections/01-introduction.tex
\section{Introduction}
\label{sec:introduction}

Autonomous systems under partial observability do not act directly on
raw sensor events.  They first turn each noisy observation into a
posterior belief, and only then does a classical planner choose what to
do next.  The difficult case is a rare observation: the evidence term is
small, classical sampling can become expensive, and any inference error
is carried forward as the next prior.

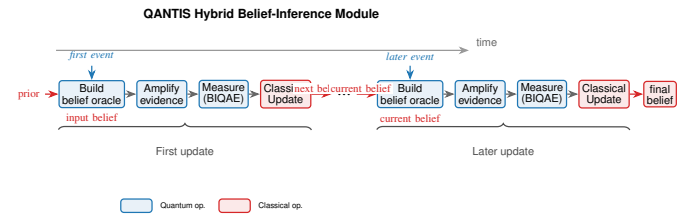
\begin{figure}[!b]
  \centering
  \resizebox{\columnwidth}{!}{\input{figures/sequential-aa-loop}}
  \caption{\QANTIS{} inference core studied in this paper.  Quantum
  amplitude amplification and boundary-aware \BIQAE{} replace the
  sampling-intensive evidence update; policy optimization, tracking,
  and \MTDA{} remain classical context rather than hardware-validated
  claims.}
  \label{fig:belief-pipeline}
\end{figure}

This paper studies one narrow hardware question.  Can a calibrated
quantum belief-update primitive be reused across several \POMDP{}
decision steps on present IBM Heron hardware without damaging the
posterior that the planner receives?  The answer is a controlled
hardware case study.  We do not claim an end-to-end autonomous-system
demonstration, a wall-clock speedup, or a validated large-state autonomy
stack.  The validated object is smaller and easier to audit: the
planner-facing posterior returned by the \QANTIS{} inference core in
Fig.~\ref{fig:belief-pipeline}.

\begin{table}[t]
\centering
\caption{Operational contract of the \QANTIS{} belief-update service.
The paper evaluates this module, not the whole autonomy stack around it.}
\label{tab:operating-contract}
\small
\setlength{\tabcolsep}{2.5pt}
\begin{tabular}{@{}p{0.24\columnwidth}p{0.68\columnwidth}@{}}
\toprule
Stage & Contract checked in this paper \\
\midrule
Input & A prior belief and observation model from the classical planner. \\
Quantum step & Estimate the rare-event evidence term using calibrated
amplification and \BIQAE{}. \\
Safety check & Compare the returned posterior with exact Bayes and verify
whether the same action would be chosen. \\
Output & An ordinary posterior distribution; downstream planning remains
classical and out of scope. \\
\bottomrule
\end{tabular}
\end{table}

The paper uses a small acronym set repeatedly: amplitude amplification
(AA) boosts rare evidence events; fixed-point AA (\FPAA{}) uses softer
phases to avoid overshoot; \BIQAE{} estimates the resulting amplitude;
and the \POMDP{} is the classical belief-and-action model.  The reason
for using AA is simple: in the rare-event regime it changes the
inference-side sampling burden from inverse-evidence sampling toward
inverse-square-root-evidence sampling~\cite{brassard_amplification_2002}.
The systems risk is also simple: a belief tracker is sequential, so a
small posterior error at one step becomes the prior at the next.  We
therefore evaluate posterior fidelity, same-trajectory amplification
controls, action agreement under a standard Tiger decision rule, and
resource accounting rather than reporting isolated amplitude gains.

\begin{table}[t]
\centering
\caption{Claim hierarchy used throughout the arXiv revision.  This
separates the primary result from controls and exploratory scaling
evidence.}
\label{tab:claim-hierarchy}
\footnotesize
\setlength{\tabcolsep}{3pt}
\begin{tabular}{@{}p{0.25\columnwidth}p{0.68\columnwidth}@{}}
\toprule
Level & Evidence used in this paper \\
\midrule
Primary claim &
Sequential Tiger \POMDP{} belief updates on Heron with posterior
Hellinger distance, matched-shot controls, and action/value checks. \\
Supporting controls &
Heron~R3 transfer, 20-step and 32-step repeated trajectories,
mitigation A/B, boundary-aware \BIQAE{}, and rare-event logical
amplification. \\
Exploratory evidence &
4-state corridor and UCGate/QSD pilots that map the scaling path but
do not establish realistic autonomous-system readiness. \\
Out of scope &
Wall-clock quantum speedup, full policy optimization on hardware, and
hardware advantage for downstream \MTDA{}. \\
\bottomrule
\end{tabular}
\end{table}

Table~\ref{tab:claim-hierarchy} is the reading key for the rest of the
paper.  The main result is the sequential two-state Tiger case study.
The longer horizons, Heron~R3 runs, rare-event envelope, and scaling
pilots are supporting evidence or operating-boundary probes.  They are
included so the reader can see where the module works today and where
the next hardware generation is still needed.

For context, \QANTIS{} v1~\cite{eker_qantis_v1_2026} validated a
broader platform around a guarded single-iterate Grover-AA Tiger update
on IBM Heron, with max Hellinger below 0.015 in the reported
8-step loop.  This revision is intentionally narrower.  It keeps the
downstream autonomy stack classical, removes the skip guard by testing
all-step \FPAA{}, calibrates the \BIQAE{} estimation layer at
near-zero and near-one amplitudes, adds action/value and resource checks, and treats longer
horizons and larger encodings as operating-boundary evidence rather than
as standalone autonomy claims.

\paragraph{Contributions}
The paper makes four focused contributions.

\begin{enumerate}
  \item \textbf{Planner-facing sequential inference.}
    \FPAA{} keeps Tiger \POMDP{} posteriors within max Hellinger
    0.009 on the 8-step run and 0.021 on the 12-step run; the revised analysis
    also reports action agreement and value loss.
  \item \textbf{Matched controls and resource accounting.}
    No-AA, guarded Grover-AA, and all-step \FPAA{} are compared on the
    same trajectory, with shot budgets, circuit counts, and missing
    queue/execute timing fields made explicit.
  \item \textbf{Boundary-aware estimation.}
    Calibrated \BIQAE{} reduces same-backend Pittsburgh boundary error
    from 0.6317 to 0.00224 at amplitude 0.01 and from 0.4890 to
    0.00773 at amplitude 0.95.
  \item \textbf{Operating boundary.}
    Heron~R3 transfer, 20-step and 32-step controls, rare-event logical
    amplification, and UCGate pilots identify circuit depth per step as
    the current coherence constraint.
\end{enumerate}

\paragraph{Paper organization}
Section~\ref{sec:background} defines the minimal notation.  Section~\ref{sec:multistep-grover}
contains the sequential hardware study, matched controls, and
decision-impact check.  Section~\ref{sec:biqae-calibration} gives the
boundary-aware \BIQAE{} calibration.  Section~\ref{sec:nisq-feasibility}
collects transfer, resource, mitigation, and scaling evidence.
Section~\ref{sec:conclusion} closes with the autonomous-systems
interpretation.

\input{sections/00-related-work-block}

%% file: figures/sequential-aa-loop.tex

\begin{tikzpicture}[
    >=Stealth,
    node distance=0.4cm and 0.25cm,
    qbox/.style={
        draw=quantum-blue, fill=quantum-blue!10, thick,
        rounded corners=2pt, minimum height=0.55cm,
        font=\scriptsize\sffamily, align=center, inner sep=2pt,
    },
    cbox/.style={
        draw=classical-red, fill=classical-red!10, thick,
        rounded corners=2pt, minimum height=0.55cm,
        font=\scriptsize\sffamily, align=center, inner sep=2pt,
    },
    obs/.style={
        font=\scriptsize\itshape, text=quantum-blue,
    },
    belief/.style={
        ->, thick, classical-red,
    },
    timeline/.style={
        ->, thick, black!60,
    },
    brace/.style={
        decorate, decoration={brace, amplitude=4pt, raise=2pt},
        thick, black!70,
    },
]

\node[font=\footnotesize\sffamily\bfseries, anchor=south]
    at (3.8, 2.72) {\QANTIS{} Hybrid Belief-Inference Module};

\node[qbox] (prep1) at (0, 1.2) {Build\\belief oracle};
\node[qbox, right=of prep1] (grover1) {Amplify\\evidence};
\node[qbox, right=of grover1] (meas1) {Measure\\[-1pt](\BIQAE)};
\node[cbox, right=of meas1] (update1) {Classical\\[-1pt]Update};

\node[obs, above=0.32cm of prep1] (o1) {first event};
\draw[->, quantum-blue, thick] (o1) -- (prep1);

\draw[timeline] (prep1) -- (grover1);
\draw[timeline] (grover1) -- (meas1);
\draw[timeline] (meas1) -- (update1);

\node[
    font=\scriptsize,
    text=classical-red,
    fill=white,
    inner sep=1pt
] (b1) at ([yshift=-0.22cm]prep1.south) {input belief};

\draw[brace] ([yshift=-0.55cm]prep1.south west) --
    node[below=5pt, font=\scriptsize\sffamily] {First update}
    ([yshift=-0.55cm]update1.south east);

\node[font=\large, right=0.45cm of update1] (dots) {...};

\node[qbox, right=0.45cm of dots] (prepT) {Build\\belief oracle};
\node[qbox, right=of prepT] (groverT) {Amplify\\evidence};
\node[qbox, right=of groverT] (measT) {Measure\\[-1pt](\BIQAE)};
\node[cbox, right=of measT] (updateT) {Classical\\[-1pt]Update};

\node[obs, above=0.32cm of prepT] (oT) {later event};
\draw[->, quantum-blue, thick] (oT) -- (prepT);

\draw[timeline] (prepT) -- (groverT);
\draw[timeline] (groverT) -- (measT);
\draw[timeline] (measT) -- (updateT);

\node[
    font=\scriptsize,
    text=classical-red,
    fill=white,
    inner sep=1pt
] (bT) at ([yshift=-0.22cm]prepT.south) {current belief};

\draw[brace] ([yshift=-0.55cm]prepT.south west) --
    node[below=5pt, font=\scriptsize\sffamily] {Later update}
    ([yshift=-0.55cm]updateT.south east);

\draw[belief] (update1.east) --
    node[above, pos=0.35, font=\scriptsize, fill=white, inner sep=1pt] {next belief}
    (dots.west);
\draw[belief] (dots.east) --
    node[above, pos=0.20, font=\scriptsize, fill=white, inner sep=1pt] {current belief}
    (prepT.west);

\node[font=\scriptsize, text=classical-red, left=0.3cm of prep1] (b0) {prior};
\draw[belief] (b0) -- (prep1.west);

\node[cbox, right=0.3cm of updateT, minimum height=0.5cm]
    (out) {final\\[-1pt]\scriptsize belief};
\draw[belief] (updateT.east) -- (out.west);

\draw[->, thick, black!40] (-0.8, 2.18) -- (8.45, 2.18)
    node[above right=-1pt and 1pt, font=\scriptsize\sffamily, black!60] {time};

\node[qbox, minimum height=0.3cm, minimum width=0.7cm]
    at (1.0, -1.3) (legq) {};
\node[font=\tiny\sffamily, right=0.05cm of legq] {Quantum op.};
\node[cbox, minimum height=0.3cm, minimum width=0.7cm]
    at (3.2, -1.3) (legc) {};
\node[font=\tiny\sffamily, right=0.05cm of legc] {Classical op.};

\end{tikzpicture}

%% file: sections/00-related-work-block.tex

\paragraph{Related Work and Positioning}
Prior work establishes the pieces around this study without closing the
sequential hardware loop.  Quantum \POMDP{} and quantum-MDP theory
clarifies the planning landscape~\cite{aaronson_qpomdp_2014,cunha_hybrid_2025,cintio_quantum_rl_2026,sqsd_pomdp_2026};
\QANTIS{} v1 put a guarded single-iterate belief update on IBM
Heron~\cite{eker_qantis_v1_2026}; and recent analyses explain
multi-step amplitude-amplification distortion in simulation or
idealized noise models~\cite{zecchi_oaa_distortion_2025,suzuki_grover_imaginary_2025,kruse_qrl_benchmark_2025}.

The estimation and systems-adjacent pieces are similarly close but not
identical to the present target.  \BIQAE{} and Bayesian amplitude
estimation provide the estimator~\cite{li_biqae_2026,ramoa_santos_bae_2025},
while boundary bias is known for iterative QAE~\cite{iqae_bias_2024}
without a hardware calibration layer for repeated belief updates.
Downstream assignment work
~\cite{stollenwerk_atm_2021,ihara_quantum_tracking_2025,pelofske_qa_vs_qaoa_2024}
is architectural context rather than the empirical focus.  The gap is
therefore narrow: sequential \POMDP{} belief updating on hardware, with
the boundary calibration needed to keep the loop stable.

Relative to those threads, this revision combines five pieces in one
hardware narrative: all-step fixed-point amplification, boundary-aware
estimation, a rare-event operating envelope, Heron~R3 transfer checks,
and direct validation of the returned posterior.  \QANTIS{} v1 supplied
the closest baseline, but it used a guard that skipped amplification
when the belief was already concentrated.  The present study removes
that guard and then adds the calibration layer needed to keep the
sequential loop readable by a planner.

%% file: sections/02-background.tex
\section{Problem Setting and Minimal Background}\label{sec:background}

This section is a reader's map, not a derivation.  Operationally, the
\QANTIS{} inference module receives a prior belief and an observation
model, performs one calibrated quantum-assisted update, and returns a
posterior distribution to the classical decision stack.  We therefore
introduce only three ingredients: the Bayes update that defines the
service contract, the quantum primitives that help in the rare-event
case, and the downstream assignment context that remains outside the
validated hardware claim.  Full derivations appear in
\QANTIS{} v1~\cite{eker_qantis_v1_2026}.

We keep notation deliberately light.  Standard \POMDP{} symbols are
introduced only where they are needed; the reader can follow the paper
as a service contract: prior in, observation model in, posterior out.
When the experiments say 8, 12, 20, or 32 steps, that number is the
decision horizon, not a transition-kernel parameter.

\subsection{Updating Belief in the Decision Loop}\label{ssec:pomdp-bg}

A \POMDP{} maintains a belief over hidden state and updates it by Bayes'
rule~\cite{kaelbling_pomdp_1998}: predict the next hidden-state
distribution, weight it by the observation likelihood, and normalize by
the evidence probability.  For the case study, that ordinary Bayes
update is the interface.  The quantum subroutine may help estimate the
rare-event evidence and the resulting posterior, but the planner still
consumes a classical probability distribution.  Online planners such as
\textsc{pomcp} pay heavily when the evidence is rare because they need
many more samples per belief node~\cite{silver_pomcp_2010}.  The
question studied in this paper is whether the same belief-update stage
can be supported by a quantum subroutine when the update must be
repeated sequentially across a decision horizon.

\subsection{Quantum Inference Primitives}\label{ssec:qc-bg}

The central quantum primitive is amplitude amplification.  A
state-preparation unitary places a small marked amplitude in the event
subspace, and Grover-style reflections boost that event using roughly the
inverse square root of the event probability rather than the inverse
probability required by direct sampling
~\cite{brassard_amplification_2002,nielsen_quantum_2010}.  In this paper
the marked event is not an arbitrary hidden state; it is the evidence
that an observation is accepted under the current belief model.  The
amplifier therefore acts directly on the inference bottleneck.

Reliable use of this primitive requires an amplitude-estimation layer.
When the target amplitude is unknown, \BIQAE{}~\cite{li_biqae_2026}
keeps a Bayesian posterior over the amplitude angle, runs a chosen
Grover depth, observes success or failure, and reweights the posterior
with the corresponding sinusoidal likelihood.  The exact update is
standard; the systems issue here is where it fails on hardware.  Because
the sequential belief loop repeatedly revisits amplitudes near zero and
one, estimator stability matters as much as amplification itself.
Section~\ref{sec:biqae-calibration} therefore develops the
boundary-aware calibration layer used by the hardware runs.

\subsection{Downstream Assignment Context}\label{ssec:mtda-bg}

The validated inference core sits upstream of association and tracking
tasks.  In the single-scan case, multi-target data association reduces
to a standard Hungarian assignment problem with cubic-time classical
solvers~\cite{kuhn_hungarian_1955,barshalom_tracking_1995}; the
multi-scan extension is NP-hard~\cite{reid_mht_1979}.  This downstream
layer is out of scope for the present paper; the QECS-relevant
hybrid-quantum assignment formulations
(annealing~\cite{stollenwerk_atm_2021,ihara_quantum_tracking_2025} and
gate-model~\cite{pelofske_qa_vs_qaoa_2024,saavedra_fpcqaoa_2025,farhi_qaoa_2014})
sit alongside the validated inference module rather than inside it.

%% file: sections/03-multistep-grover.tex

\section{\QANTIS{} Belief-Update Service on Heron}
\label{sec:multistep-grover}

This section is the core hardware story.  \QANTIS{} v1 already showed
that a guarded amplified Tiger belief update can run on IBM
Heron~\cite{eker_qantis_v1_2026}.  The question here is more operational:
can the quantum step be treated as a reusable service that runs at every
listen update, returns a posterior to the classical planner, and stays
stable when the belief becomes very concentrated?

The section is organized in a result-first order.  We first define what
the hardware posterior means, because that is the product interface.
We then compare no amplification, guarded Grover-\textsc{aa}, and
all-step \FPAA{} on the same Tiger trajectory.  Finally, we ask the
planner-facing question: do those posterior differences change the
immediate Tiger action or value?

\subsection{Inference Core and Error Model}
\label{ssec:error-model}

In \QANTIS{} v1, a single-iterate Grover-\textsc{aa} update applied to
a Tiger \POMDP{} belief oracle on IBM Heron~R2 yielded a 5.1-fold
amplification at ISA~15--18 and an 8-step closed-loop run with maximum
Hellinger below 0.015~\cite{eker_qantis_v1_2026}, using an adaptive
skip guard so that amplification was applied only when beneficial.
The present section keeps the same Tiger unit cell and the same Heron
backends but removes that guard: amplification is applied at every
listen step, which is the regime where boundary-driven instabilities of
the estimation layer become operationally visible.

Prior literature does not yet answer that empirical question.  Zecchi
et al.~\cite{zecchi_oaa_distortion_2025} derive distortion bounds for
iterated oblivious amplitude amplification (\OAA{}) without hardware
validation.  Cunha et al.~\cite{cunha_hybrid_2025} propose a quantum
Bayesian reinforcement-learning agent (\QBRL{}) in simulation, while
Suzuki et al.~\cite{suzuki_grover_imaginary_2025} analyze Grover-based
imaginary-time evolution in idealized noise models.  What remains open
is the decision-loop question that matters operationally: how cumulative
gate noise, shot noise, and amplification distortion alter posterior
accuracy when amplitude amplification is used repeatedly inside a
\POMDP{} belief tracker.

Operationally, each time step builds a belief oracle whose measurement
statistics encode the current belief.  The amplifier then marks the
event ``this observation is accepted under the current sensor model''
rather than marking a hidden state directly.  We use two phase choices:
a guarded Grover-\textsc{aa} baseline with the standard
Brassard--Hoyer--Mosca reflections~\cite{brassard_amplification_2002},
and an all-step fixed-point trajectory with the softer Yoder--Low--Chuang
phase schedule~\cite{yoder_fpaa_2014}.  The softer phase is the practical
point: it reduces overshoot when the belief is already concentrated, so
the service can amplify every listen step instead of relying on a skip
guard.

\paragraph{What ``hardware posterior'' means}
The hardware output of one step is the \BIQAE{}-estimated evidence
amplitude, converted back into the evidence probability for the observed
growl, and then used in the ordinary Bayes update.  The reported
posterior is therefore the planner-facing posterior returned by the
service.  The conditional accepted-shot histogram is a
hardware-credibility cross-check, not the primary extraction path.  The
exact Bayes posterior is used only as the reference for Hellinger
distance.

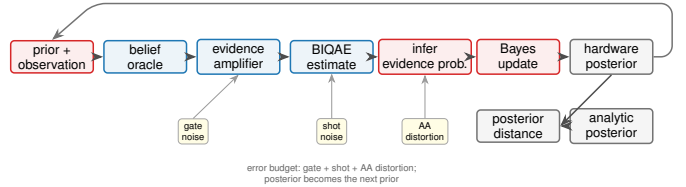
\begin{figure}[t]
  \centering
  \resizebox{\columnwidth}{!}{\input{figures/hardware-posterior-error}}
  \caption{Hardware posterior extraction and per-step error budget.}
  \label{fig:hardware-posterior-error}
\end{figure}

Three noise terms matter operationally
(Fig.~\ref{fig:hardware-posterior-error}): gate noise accumulated by
the ISA circuit, shot noise from the measurement budget, and distortion
from imperfect amplification.  Since each step posterior becomes
the prior at the next step, these errors can compound across the loop;
informative observations can also contract part of the error under the
Bayes update.  We report Hellinger distance between the hardware-derived
posterior and the exact Bayes posterior because it is bounded, symmetric,
and directly interpretable as a posterior distance.  Shot allocation is
likewise treated as part of the service: the adaptive v1 allocation
moves about 15\% of the budget toward the
early and ambiguous middle steps where shot noise is most consequential.

\subsection{Tiger Case-Study Protocol}
\label{ssec:tiger-protocol}

The Tiger \POMDP{} serves as the unit cell of the case study: small
enough to execute repeatedly on present hardware, but still rich enough
to contain the full inference mechanism of interest.  The model has
two hidden states, three actions (listen, open-left, open-right), and
85\% observation accuracy.  For a sequence of listen actions, the exact
posterior after a pattern of left and right growls is available in
closed form, which makes hardware--theory comparison precise rather than
heuristic.

Starting from a uniform prior, we execute a fixed sequence of decision
steps consisting of listen updates and occasional reset actions.  At
each listen step we construct the belief oracle for the current prior and
sensor model, apply one amplification iterate, estimate the amplified
success amplitude with \BIQAE{}~\cite{li_biqae_2026}, convert that
estimate back into the evidence probability, and use it in the analytic
Bayes update before feeding the updated belief into the next step.  Reset actions bypass
amplitude amplification and return the Tiger belief to near-uniform, so
the horizon counts decision steps rather than listen-only observations.
The oracle width remains fixed at two qubits throughout
the sequence; only the encoded rotation angles change as the belief
state evolves.

\paragraph{Service loop}
The implementation follows the same auditable loop at every decision
step.  The run first fixes the per-step shot schedule, either uniform or
the adaptive v1 allocation.  A reset event simply restores the near-uniform
Tiger belief.  A listen event builds the current belief oracle, applies
the selected one-step amplifier, estimates the amplified acceptance
amplitude with \BIQAE{}, converts the amplifier output back into the
evidence probability, and feeds that scalar into the Bayes update.  The
reported audit output is the final belief together with the per-step
Hellinger distance between the hardware-derived and exact Bayes
posteriors.

\subsection{Sequential Hardware Results}
\label{ssec:hw-results}

The main-campaign sequential experiments were executed on IBM
Heron~R2 (IBM Kingston, 156 qubits, \EPLG{}
around 0.002); supplementary transfer runs on Heron~R3
are reported in Section~\ref{ssec:heron-r3-transfer}.  Per-step circuits compile to ISA
depths of 15--18 two-qubit gates (single-iterate Grover: 15--18; Tiger
minimal oracle: 12--13), using Qiskit~2.3 transpilation at
optimization level~3.  Shot budgets are reported per-step rather than
as a global total: the 4-step baseline uses 10,000 shots per step, and
the 8-step guarded Grover-\textsc{aa} run uses the v1 adaptive schedule
between 8,000 and 16,000 shots per step, with 10,000 shots on average.
The all-step \FPAA{} run uses 32,768 shots per step, approximately three
times the
guarded-Grover per-step budget.  The \FPAA{} comparison in
Table~\ref{tab:fpaa-comparison} is therefore a same-trajectory
stability comparison, not a strictly budget-matched accuracy
comparison; we discuss this distinction explicitly there.

We probe two belief trajectories with distinct dynamics.  The
4-step baseline listens twice to left growls, resets, and then listens
once more to a left growl.  This short trajectory combines two
informative observations, a near-uniform reset, and a final
re-concentration of the posterior.  The 8-step stress test uses the
same opening pattern but adds two right growls and two final left
growls, so the belief passes through ambiguous as well as informative
states.  This second sequence is the more relevant stress test because
it repeatedly moves the inference core between regimes where
amplification is useful or risky.

Table~\ref{tab:multistep-results} is the audit trail for the first
hardware comparison.  The important reading point is not any single row:
the no-AA and guarded Grover rows use the same backend and, for the
8-step case, the same decision path.  That alignment lets us interpret
amplification as a change to the inference mechanism rather than a
change of task instance.

\begin{table}[t]
\centering
\caption{Per-step posterior accuracy on IBM Kingston (Heron~R2).
  Bold entries mark the per-sequence maximum Hellinger distance.}
\label{tab:multistep-results}
\resizebox{\columnwidth}{!}{%
\begin{tabular}{@{}c c c c c@{}}
\toprule
Step & Event & Grover & Amp. & Hellinger \\
\midrule
\multicolumn{5}{@{}l}{4-step baseline, no AA, on IBM Kingston} \\
\addlinespace[2pt]
0 & listen(0) & --- & --- & 0.0030 \\
1 & listen(0) & --- & --- & 0.0040 \\
2 & open-right & reset & --- & 0.0003\textsuperscript{a} \\
3 & listen(0) & --- & --- & 0.0126 \\
\midrule
\multicolumn{5}{@{}l}{8-step no-AA baseline on IBM Kingston} \\
\addlinespace[2pt]
0 & listen(0) & --- & --- & 0.0123 \\
1 & listen(0) & --- & --- & 0.0019 \\
2 & open-right & reset & --- & 0.0195\textsuperscript{a} \\
3 & listen(0) & --- & --- & 0.0120 \\
4 & listen(1) & --- & --- & 0.0222 \\
5 & listen(1) & --- & --- & 0.0172 \\
6 & listen(0) & --- & --- & \textbf{0.0286} \\
7 & listen(0) & --- & --- & 0.0078 \\
\midrule
\multicolumn{5}{@{}l}{8-step Grover-AA with guard on IBM Kingston} \\
\addlinespace[2pt]
0 & listen(0) & \cmark & 1.02x & 0.0251 \\
1 & listen(0) & \xmark\ (skip) & --- & 0.0167 \\
2 & open-right & reset & --- & 0.0076\textsuperscript{a} \\
3 & listen(0) & \cmark & 1.07x & 0.0267 \\
4 & listen(1) & \cmark & 3.38x & 0.0048 \\
5 & listen(1) & \xmark\ (skip) & --- & 0.0020 \\
6 & listen(0) & \cmark & 4.17x & 0.0005 \\
7 & listen(0) & \cmark & 1.29x & \textbf{0.0274} \\
\bottomrule
\multicolumn{5}{@{}l}{\footnotesize \textsuperscript{a}Open-right action resets belief to near-uniform.} \\
\multicolumn{5}{@{}l}{\footnotesize No-AA 8-step: max Hellinger 0.0286, mean 0.0152.} \\
\multicolumn{5}{@{}l}{\footnotesize Grover-AA 8-step: 5/7 listen steps; mean amp. 2.18x; max Hellinger 0.0274, mean 0.0139.} \\
\multicolumn{5}{@{}l}{\footnotesize Rare-row Hellinger drops: 4.6x at the first reversal and 57x at the later reversal.}
\end{tabular}}
\end{table}

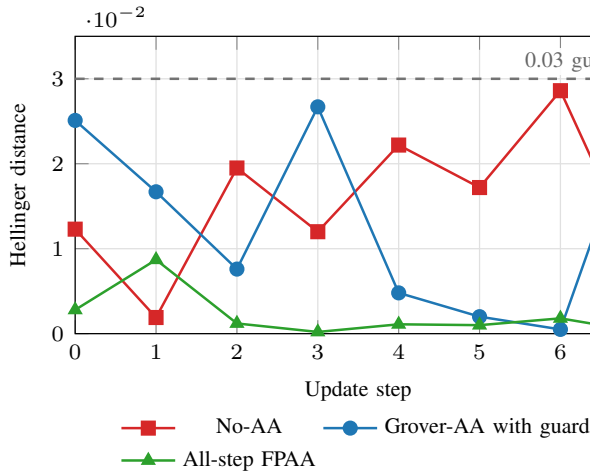
\begin{figure}[t]
  \centering
  \resizebox{\columnwidth}{!}{\input{figures/multistep-hardware-results}}
  \caption{Per-step Hellinger distance on the shared 8-step
  decision-step path.  The revised arXiv figure plots No-AA, guarded
  Grover-AA, and all-step \FPAA{} together instead of forcing the
  comparison into a dense table.}
  \label{fig:multistep-hardware}
\end{figure}

Table~\ref{tab:multistep-results} is the audit view; the text can be
read more simply.  The guarded Grover run stays inside the engineering
band, the exact MAP decision is unchanged, and the largest benefit
appears exactly where the service needs help: rare observations rather
than near-uniform updates.  The yield improvement in those moderate-rare
Tiger steps is about 3.6--4.2x, but that is a sample-efficiency result,
not a wall-clock latency claim.  Compilation, queueing, and runtime
overheads are therefore kept outside the claim
boundary~\cite{brassard_amplification_2002,zalka_grover_optimal_1999,legall_dequantizing_shortpath_2026}.
The ultra-rare regime is treated separately in
Section~\ref{sssec:fpaa-boundary}, where it is labeled explicitly as a
logical sample-complexity envelope.

\subsection{Fixed-Point Amplitude Amplification for Stable Belief Tracking}
\label{ssec:fpaa}

The next step is to remove the decision-rule fragility that still
remains in standard Grover-AA.  Standard Grover reflections are well
suited to very small success amplitudes, but can overshoot once the
posterior is already moderately concentrated.  In a sequential belief
loop, that means the controller must decide when to apply AA and when to
skip it.  Fixed-point amplitude amplification
(\FPAA{})~\cite{yoder_fpaa_2014} addresses exactly that systems issue
by replacing hard phase flips with softer fixed-point rotations, so the
update remains monotone rather than oscillatory.

We compare three inference policies on the same 8-step path used in
Table~\ref{tab:multistep-results}: no amplification (No-AA), guarded
Grover-AA, and \FPAA{} applied at every listen step.  No-AA is the
classical sequential Bayes baseline: identical backend and observation
sequence, but no amplitude amplification.  The original conference
draft blurred two issues here: the 32\,768-shot \FPAA{} headline
showed best posterior fidelity, while the budget-relevant question is
whether the algorithmic choice survives a comparable budget.  The
arXiv revision therefore reports both the 32\,768-shot headline and
the 10\,000-shot matched-shot \FPAA{} control in
Table~\ref{tab:fpaa-comparison}.  The guarded Grover row is
budget-aligned but not strictly constant-shot, because its recorded run
uses the v1 adaptive allocation, between 8k and 16k shots/step with
about 10k shots/step on average.  We therefore do not call the table a fully
budget-matched proof; it is a same-trajectory hardware control that
exposes the remaining constant-shot Grover rerun needed for a future
claim.

\begin{table}[t]
\centering
\caption{Same-trajectory amplification controls on IBM Kingston,
  8-step path.  The 10k \FPAA{} row is the matched-shot hardware control;
  the Grover row uses the recorded adaptive v1 allocation and is
  therefore budget-aligned rather than strictly constant-shot.}
\label{tab:fpaa-comparison}
\resizebox{\columnwidth}{!}{%
\begin{tabular}{@{}l c c c c@{}}
\toprule
Metric & No-AA & Grover-AA & \FPAA{} control & \textbf{\FPAA{} headline} \\
\midrule
Shots/step     & 10k & 8k--16k & 10k & 32\,768 \\
Steps applied  & 0/7 & 5/7 & \textbf{7/7} & \textbf{7/7} \\
Max Hellinger  & 0.029 & 0.027 & 0.033 & \textbf{0.009} \\
Mean Hellinger & 0.015 & 0.014 & 0.019 & \textbf{0.004} \\
\bottomrule
\end{tabular}
}
\end{table}

\begin{table}[t]
\centering
\caption{Longer-horizon all-step \FPAA{} stability checks.  These rows
extend the operating envelope; they do not broaden the primary 8-step
and 12-step claim.}
\label{tab:long-horizon-fpaa}
\footnotesize
\setlength{\tabcolsep}{3pt}
\begin{tabular}{@{}l c c c@{}}
\toprule
Run & Listen steps amplified & Max Hellinger & Mean Hellinger \\
\midrule
12-step Fez repeat & 9/9 & 0.021 & 0.009 \\
12-step Fez repeat & 9/9 & 0.010 & 0.004 \\
20-step Fez repeat & 15/15 & 0.025 & 0.007 \\
20-step Fez repeat & 15/15 & 0.021 & 0.005 \\
32-step Fez run & 25/25 & 0.023 & 0.006 \\
\bottomrule
\end{tabular}
\end{table}

All-step \FPAA{} is therefore best read as a stability result.  It
amplifies every listen step and gives the best posterior fidelity in the
high-shot headline run, while the matched-shot ablation exposes the
remaining budget-matched Grover control that a future claim would need.
A Fez repeat run and a Kingston mitigation A/B run land in the same
stability band, so we do not treat the result as a one-shot calibration
accident.
The per-step pattern is plotted in Fig.~\ref{fig:multistep-hardware}:
the all-step \FPAA{} curve stays near the bottom of the axis for most
steps, with two Kingston executions bracketing the headline 0.009.

Table~\ref{tab:long-horizon-fpaa} keeps the longer-horizon details out
of the prose.  The reading point is that the service does not immediately
collapse when the posterior is fed back for many more updates; the
primary claim remains narrower.

\paragraph{Decision impact}
For a planner, a small posterior error matters only if it changes the
action.  We therefore translate the posterior into the standard Tiger
immediate-reward rule: listen is cheap, opening the safe door is
rewarding, and opening the tiger door is costly.  The exact Bayes
posterior and the hardware posterior are passed through the same rule at
every step, and any disagreement is scored as value loss under the exact
posterior.  In this rule the planner opens right above
90\% belief that the tiger is left, opens left below 10\%, and listens
between those thresholds.  Table~\ref{tab:decision-impact} gives the outcome: across
the matched-shot and extended-horizon checks, the hardware posterior
preserves the planner-facing action.  The remaining deviations are
posterior-fidelity errors, not action flips under this decision rule.

\begin{table}[t]
\centering
\caption{Decision-impact check using the standard Tiger immediate
reward rule.  Value loss is computed under the exact Bayes posterior
when the hardware posterior selects a different action.}
\label{tab:decision-impact}
\small
\setlength{\tabcolsep}{3pt}
\resizebox{\columnwidth}{!}{%
\begin{tabular}{@{}l c c c c@{}}
\toprule
Trajectory & Steps & Max Hellinger & Action agree. & Cum. value loss \\
\midrule
8-step matched-shot \FPAA{} & 8  & 0.0326 & 8/8   & 0.000 \\
20-step \FPAA{} control     & 20 & 0.0343 & 20/20 & 0.000 \\
32-step \FPAA{} control     & 32 & 0.0228 & 32/32 & 0.000 \\
\bottomrule
\end{tabular}
}
\end{table}

\subsubsection{Rare-Observation Envelope}
\label{sssec:fpaa-boundary}

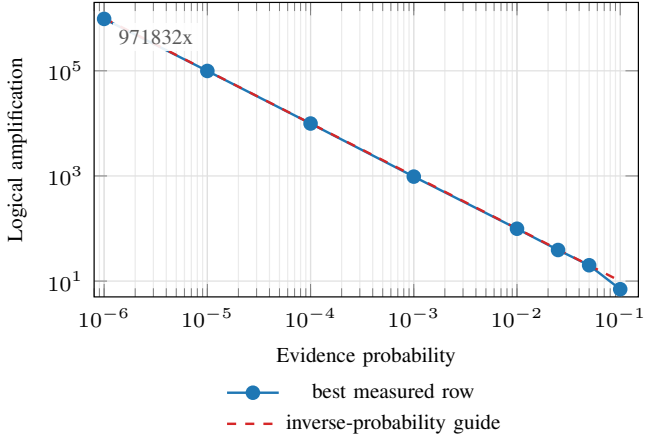
\begin{figure}[t]
  \centering
  \resizebox{\columnwidth}{!}{\input{figures/rare-event-envelope}}
  \caption{Rare-observation envelope for Grover--Brassard iterates on a
  one-qubit logical oracle.  The plot reports the best measured row at
  each evidence probability; full per-depth rows, raw counts, and backend
  manifests remain in the artifact bundle.}
  \label{fig:rare-event-envelope}
\end{figure}

Yoder--Low--Chuang~\cite{yoder_fpaa_2014} gives a bounded fixed-point
schedule once a lower bound on the event amplitude is known.  Our
probe-then-amplify deployment is related to the nested amplification
protocol of \cite{demmler_nested_aa_2026} (binary knapsack GAS); we keep
the bounded-length guarantee and replace the GAS outer loop with a
\POMDP{} belief-update oracle.  We verify the prediction under
superconducting noise via a one-qubit Grover-Brassard amplifier,
primarily on IBM Kingston
(Heron~R2), with one IBM Aachen Heron~R3 crossover row
(Fig.~\ref{fig:rare-event-envelope}).

The envelope in Fig.~\ref{fig:rare-event-envelope} should be read as an
operating map, not as a deep-circuit fidelity claim.  Its role is to
answer a simple systems question: if the evidence event is extremely
rare, can the hardware run keep the accepted-event probability near the
amplified target long enough for the belief-update service to use it?
At the rarest tested point, a one-in-a-million evidence event, Kingston
R2 measures 0.9718 acceptance against the 0.9714 analytic prediction.
We therefore report that row as 971832x \emph{logical} amplification.

The caveat is deliberately kept next to the claim.  After transpilation
at the runner's default optimization level, the source circuit collapses
to at most six single-qubit basis gates.  A separate barrier-fenced
depth sweep, with raw identifiers kept in the artifact manifest,
confirms the boundary: merged circuits track the logical prediction,
whereas deliberately unmerged deeper circuits mark the physical-depth
frontier.  In other words, the rare-event figure validates the
sample-complexity envelope that the service wants to exploit; it does
not by itself validate long, deep rare-event circuits.

The remaining rows are audit checks.  They show that the bounded-length
schedule is not just fitting one backend snapshot: the next rare row
reaches 99402x logical amplification, the 0.0001 row reproduces across
backends within 0.15\%, and Fez cross-validation stays within 1.62\% of
theory, including the 0.01 row at 99x amplification.

\paragraph{Two related but distinct circuit families}
For clarity in the rest of the paper we use ``all-step \FPAA{}'' for
the Tiger sequential loop with phase-shifted reflections
and a single iterate per listen step, as used in
Section~\ref{ssec:fpaa} and Table~\ref{tab:fpaa-comparison}; we use
``bounded-length Grover--Brassard amplification'' for the one-qubit
rare-event envelope of Fig.~\ref{fig:rare-event-envelope}, where the
reflections remain standard Brassard--Hoyer--Mosca reflections and the
Yoder--Low--Chuang lower-bound condition is used only to clamp the
iteration count.  These are related but distinct circuit families:
the Tiger \FPAA{} loop demonstrates sequential posterior preservation,
while the bounded-length envelope validates the rare-event acceptance
regime on a one-qubit oracle.

Operationally, the bounded schedule raises the measured
accepted-event probability close to one with a number of Grover iterates
that follows inverse-square-root evidence scaling.  A crude
Monte-Carlo baseline~\cite{asmussen_glynn_2007} would need on the order
of ten billion samples for 1\% accuracy at one-in-a-million evidence
probability, a regime that appears in autonomous-vehicle rare-event
simulation~\cite{chen_av_rare_event_2026}.  The companion software
routine clamps the iteration count from a lower bound on the event
amplitude; despite the historical name,
Fig.~\ref{fig:rare-event-envelope} circuits use Grover--Brassard
reflections.  Near-certain amplitudes motivate the complementary guard
and boundary-aware \BIQAE{} calibration: when the event probability is
unknown, a fixed Grover length can over-rotate instead of amplifying
useful evidence.

A 24-row Kingston boundary sweep confirms the adaptive-guard regime:
21 of 24 rows stay below Hellinger 0.02.  The longer Fez repeats in
Table~\ref{tab:long-horizon-fpaa} then show that the sequential service
stays inside the same operating band when the horizon is extended.

The sequential Hellinger numbers depend on stable amplitude
estimation at per-step posterior boundaries: listen updates can drive
the estimated success probability toward zero or one, where standard
\BIQAE{} becomes ill-conditioned.  Section~\ref{sec:biqae-calibration}
formalises that failure mode and introduces the two-phase boundary
calibration used by the runs above.

%% file: figures/hardware-posterior-error.tex
\begin{tikzpicture}[
    >=Stealth,
    font=\sffamily,
    qstage/.style={
        draw=quantum-blue, fill=quantum-blue!8, thick,
        rounded corners=2pt, minimum width=1.55cm, minimum height=0.55cm,
        font=\scriptsize\sffamily, align=center, inner sep=2pt
    },
    cstage/.style={
        draw=classical-red, fill=classical-red!8, thick,
        rounded corners=2pt, minimum width=1.55cm, minimum height=0.55cm,
        font=\scriptsize\sffamily, align=center, inner sep=2pt
    },
    outstage/.style={
        draw=black!55, fill=black!4, thick,
        rounded corners=2pt, minimum width=1.55cm, minimum height=0.55cm,
        font=\scriptsize\sffamily, align=center, inner sep=2pt
    },
    err/.style={
        draw=black!35, fill=yellow!12,
        rounded corners=2pt, font=\tiny\sffamily, align=center,
        inner sep=2pt
    },
    arr/.style={->, thick, black!70},
]

\node[cstage] (prior) at (0, 1.25) {prior +\\observation};
\node[qstage] (oracle) at (1.75, 1.25) {belief\\oracle};
\node[qstage] (amp) at (3.5, 1.25) {evidence\\amplifier};
\node[qstage] (biqae) at (5.25, 1.25) {\BIQAE{}\\estimate};
\node[cstage] (invert) at (7.0, 1.25) {infer\\evidence prob.};
\node[cstage] (bayes) at (8.75, 1.25) {Bayes\\update};
\node[outstage] (posterior) at (10.5, 1.25) {hardware\\posterior};

\draw[arr] (prior) -- (oracle);
\draw[arr] (oracle) -- (amp);
\draw[arr] (amp) -- (biqae);
\draw[arr] (biqae) -- (invert);
\draw[arr] (invert) -- (bayes);
\draw[arr] (bayes) -- (posterior);

\node[outstage, minimum width=1.55cm] (truth) at (10.5, -0.05) {analytic\\posterior};
\node[outstage, minimum width=1.55cm] (hell) at (8.75, -0.05) {posterior\\distance};
\draw[arr] (posterior.south) -- (hell.east);
\draw[arr] (truth.west) -- (hell.east);

\node[err] (gate) at (2.65, -0.15) {gate\\noise};
\node[err] (shot) at (5.25, -0.15) {shot\\noise};
\node[err] (dist) at (7.0, -0.15) {AA\\distortion};
\draw[->, black!40] (gate.north) -- (amp.south);
\draw[->, black!40] (shot.north) -- (biqae.south);
\draw[->, black!40] (dist.north) -- (invert.south);

\node[font=\tiny\sffamily, align=center, text=black!60] at (5.25, -0.95)
  {error budget: gate + shot + AA distortion;\\posterior becomes the next prior};
\draw[arr, rounded corners=6pt, black!55]
  (posterior.east) -- ++(0.35,0) -- ++(0,1.0) -- ++(-10.85,0) -- (prior.north);

\end{tikzpicture}

%% file: figures/multistep-hardware-results.tex
\begin{tikzpicture}
\begin{axis}[
    width=7.9cm,
    height=4.9cm,
    xmin=0, xmax=7,
    ymin=0, ymax=0.035,
    xtick={0,1,2,3,4,5,6,7},
    xlabel={Update step},
    ylabel={Hellinger distance},
    grid=both,
    grid style={black!8},
    major grid style={black!12},
    legend style={
        at={(0.5,-0.24)},
        anchor=north,
        legend columns=2,
        draw=none,
        fill=none,
        font=\scriptsize
    },
    tick label style={font=\scriptsize},
    label style={font=\scriptsize},
]
\addplot[
    color=classical-red,
    mark=square*,
    thick
] coordinates {
    (0,0.0123) (1,0.0019) (2,0.0195) (3,0.0120)
    (4,0.0222) (5,0.0172) (6,0.0286) (7,0.0078)
};
\addlegendentry{No-AA}

\addplot[
    color=quantum-blue,
    mark=*,
    thick
] coordinates {
    (0,0.0251) (1,0.0167) (2,0.0076) (3,0.0267)
    (4,0.0048) (5,0.0020) (6,0.0005) (7,0.0274)
};
\addlegendentry{Grover-AA with guard}

\addplot[
    color=dwave-green,
    mark=triangle*,
    thick
] coordinates {
    (0,0.0028) (1,0.0087) (2,0.0012) (3,0.0002)
    (4,0.0011) (5,0.0010) (6,0.0018) (7,0.0002)
};
\addlegendentry{All-step FPAA}

\addplot[
    color=black!55,
    dashed,
    thick
] coordinates {(0,0.03) (7,0.03)};
\node[
    anchor=north east,
    font=\scriptsize,
    text=black!60,
    fill=white,
    fill opacity=0.9,
    text opacity=1,
    inner sep=1pt
]
    at (axis description cs:0.97,0.96) {0.03 guide};
\end{axis}
\end{tikzpicture}

%% file: figures/rare-event-envelope.tex
\begin{tikzpicture}
\begin{axis}[
    width=7.9cm,
    height=5.0cm,
    xmode=log,
    ymode=log,
    xmin=8e-7, xmax=1.5e-1,
    ymin=5, ymax=2e6,
    xlabel={Evidence probability},
    ylabel={Logical amplification},
    grid=both,
    grid style={black!8},
    major grid style={black!12},
    tick label style={font=\scriptsize},
    label style={font=\scriptsize},
    legend style={
        at={(0.5,-0.25)},
        anchor=north,
        draw=none,
        fill=none,
        font=\scriptsize
    },
]
\addplot[
    color=quantum-blue,
    mark=*,
    thick
] coordinates {
    (1e-6,971832)
    (1e-5,99402)
    (1e-4,9935)
    (1e-3,971)
    (1e-2,99)
    (2.5e-2,39)
    (5e-2,20)
    (1e-1,7)
};
\addlegendentry{best measured row}

\addplot[
    color=classical-red,
    dashed,
    thick
] coordinates {
    (1e-6,1000000)
    (1e-5,100000)
    (1e-4,10000)
    (1e-3,1000)
    (1e-2,100)
    (1e-1,10)
};
\addlegendentry{inverse-probability guide}

\node[
    anchor=north west,
    font=\scriptsize,
    text=black!65,
    fill=white,
    fill opacity=0.9,
    text opacity=1,
    inner sep=2pt
]
    at (axis cs:1.2e-6,8e5) {971832x};
\end{axis}
\end{tikzpicture}

%% file: sections/04-biqae-calibration.tex
\section{Boundary-Aware BIQAE Calibration}
\label{sec:biqae-calibration}

Because the sequential belief loop depends on repeated amplitude
estimation, calibration is not a side contribution; it is the guardrail
that keeps the inference service trustworthy on real hardware.  The
failure mode is easy to state: when the true amplitude is very close to
zero or one, the default estimator wastes probability mass in the wrong
part of the interval and can drift toward the interior.  In a belief
tracker, that bad estimate becomes the next prior.  This section adds a
small calibration step that asks, before the expensive estimation loop,
whether the amplitude is near zero, near one, or safely interior.

\subsection{Why Boundary Calibration Matters for Sequential Inference}
\label{subsec:boundary-error}

In \QANTIS{} v1~\cite{eker_qantis_v1_2026} we observed a pronounced
U-shaped error profile when running \BIQAE{}~\cite{li_biqae_2026}
across the full amplitude range: estimation accuracy degrades sharply
as the true amplitude approaches either boundary, while interior
amplitudes remain well behaved.  Neither \BIQAE{} nor
BAE~\cite{ramoa_santos_bae_2025} directly addresses this regime, because both
default to generic priors.  Miyamoto~\cite{iqae_bias_2024}
analyzes related bias effects near the boundaries, but does not provide
a hardware calibration strategy for the sequential inference setting
studied here.

The issue is informational rather than cosmetic.  When the amplitude is
close to zero or one, a diffuse prior spreads probability mass across the full unit
interval and places only a small fraction near the true value.  The
posterior then asks for deeper Grover circuits to recover, which is
precisely the wrong behavior on noisy hardware.  The calibration below
is therefore a routing decision: spend a tiny number of shallow shots to
choose a sensible prior, then let the ordinary \BIQAE{} machinery do the
fine estimate.

\begin{figure}[t]
  \centering
  \resizebox{\columnwidth}{!}{\input{figures/biqae-calibration-compact}}
  \caption{Two-phase boundary-aware \BIQAE{} calibration.}
  \label{fig:biqae-calibration-flow}
\end{figure}
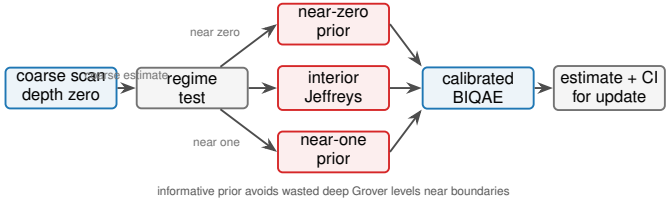

\subsection{Two-Phase Calibration Protocol}
\label{subsec:two-phase}

We therefore use a lightweight two-phase protocol that injects an
informative prior before the main \BIQAE{} loop, while preserving the
same Bayesian update structure (Fig.~\ref{fig:biqae-calibration-flow}).
The protocol is deliberately a routing layer rather than a new estimator:
use a shallow measurement to choose the prior family, then let the
ordinary \BIQAE{} grid update produce the posterior mean and the
95\% credible interval.

Phase~1 performs a shallow coarse scan using only state preparation and
measurement, with no Grover amplification.  The observed success
frequency determines whether the amplitude is near zero, near one, or
interior.  Near-zero scans receive a Beta prior concentrated near zero;
near-one scans receive the mirrored near-one prior; interior scans use a
Jeffreys prior.  Phase~2 then runs the ordinary \BIQAE{} grid update
with this boundary-aware initial prior replacing the default uniform
prior.

The protocol is motivated by information geometry as much as by
hardware pragmatics.  In the same-backend Pittsburgh paired boundary
run at amplitude 0.01, the coarse scan returns 0.025 and therefore
initializes a strongly near-zero prior, which places substantially
more mass near zero than the uniform prior.  That head start saves
2--3 Grover-depth levels in practice, which is precisely where noise
starts to dominate on current devices.  As long as the coarse estimate
is within a factor of two of the true amplitude, the posterior still
converges to the same credible-interval width with at most one extra
Grover level, which is consistent with the behavior analyzed
in~\cite{li_biqae_2026}.

The estimation layer also benefits from a noise-aware likelihood.
Standard \BIQAE{} uses an ideal amplified sinusoid, which ignores
depolarizing attenuation.  Following
Ramoa \& Santos~\cite{ramoa_santos_bae_2025}, we instead mix that ideal
sinusoid with a uniform readout floor using a visibility parameter
between zero and one.  Because this visibility is not jointly
identifiable with the unknown amplitude from a single coarse scan, we
plug in a session-level estimate obtained from short calibration circuits
at known amplitudes, including the interior row of
Table~\ref{tab:biqae-results} run in the same scheduling window.  The
resulting visibility is therefore a hardware-state plug-in rather than a
quantity inferred from the unknown-amplitude run, and we do not claim
joint identifiability.  This
modification matters most when boundary amplitudes are measured with
shallow circuits that are already noticeably attenuated by hardware
noise.

\subsection{Hardware Calibration Results}
\label{subsec:biqae-results}

The main calibration evidence in this section now comes from same-day,
same-backend paired runs on IBM Pittsburgh at the two extreme
boundaries, amplitudes 0.01 and 0.95.  We retain the earlier Heron~R2 rows
at amplitudes 0.10, 0.50, and 0.90 only as supporting context for the
moderate-boundary and interior regimes.  All calibrated runs use the
  two-phase protocol of Fig.~\ref{fig:biqae-calibration-flow} with
200 coarse shots and boundary threshold 0.1.

\begin{table}[t]
\centering
\caption{Boundary-aware \BIQAE{} calibration.  Shot acct.\ =
Phase~1~+~Phase~2 shots.}
\label{tab:biqae-results}
\setlength{\tabcolsep}{3pt}
\small
\resizebox{\columnwidth}{!}{%
\begin{tabular}{@{}cllcccc@{}}
\toprule
Amplitude & Prior & Backend & Estimate & Abs. err. & CI width & Shot acct. \\
\midrule
0.01 & Baseline & IBM Pittsburgh & 0.6417 & 0.6317 & 0.8255 & 0+2\,400 \\
     & Calib.  & IBM Pittsburgh & 0.0122 & \textbf{0.00224} & 0.00696 & 200+300 \\
\addlinespace
0.10 & Baseline & IBM Fez        & 0.446 & 0.346 & 0.861 & 0+2\,400 \\
     & Calib.  & IBM Kingston    & 0.109 & \textbf{0.009} & 0.019 & 200+300 \\
\addlinespace
0.50 & Baseline & IBM Kingston   & 0.507 & 0.007 & 0.008 & 0+2\,400 \\
\addlinespace
0.90 & Baseline & IBM Torino     & 0.556 & 0.344 & 0.861 & 0+2\,400 \\
     & Calib.  & IBM Fez         & 0.887 & \textbf{0.013} & 0.025 & 200+300 \\
\addlinespace
0.95 & Baseline & IBM Pittsburgh & 0.4610 & 0.4890 & 0.8684 & 0+2\,400 \\
     & Calib.  & IBM Pittsburgh & 0.9423 & \textbf{0.00773} & 0.0183 & 200+300 \\
\bottomrule
\end{tabular}
}
\end{table}

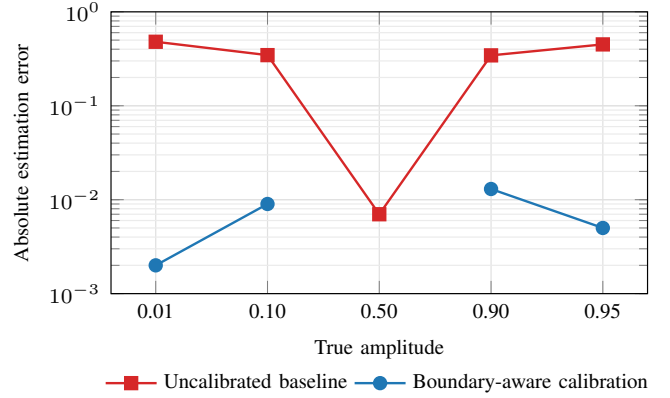
\begin{figure}[t]
  \centering
  \resizebox{\columnwidth}{!}{\input{figures/biqae-boundary-results}}
  \caption{Boundary behavior of \BIQAE{} on hardware.  Calibration
  collapses the near-zero and near-one error spikes.}
  \label{fig:biqae-boundary}
\end{figure}

Table~\ref{tab:biqae-results} shows the operational effect.  At the two
extreme Pittsburgh boundary points, the default estimator drifts toward
the interior; the calibrated estimator stays near the true amplitude and
finishes after one fine \BIQAE{} iteration.  The moderate-boundary rows
show the same pattern, while the interior point already behaves well
without calibration.  That contrast is important: the protocol is not
rebranding all of \BIQAE{} as better; it fixes the boundary regime that
the sequential belief loop repeatedly encounters.

The coarse scan selected the correct near-zero, near-one, or interior
regime in every tested case.  Its overhead is deliberately small:
200 coarse shots add a shallow state-preparation layer and no extra
Grover amplification.  The table
keeps the exact amplitudes, errors, and shot accounting for auditability;
the reader-facing takeaway is that calibration turns a fragile boundary
estimator into a stable service step.

The noise-aware likelihood is most visible in the 12-step \FPAA{} run
on IBM Fez.  With the visibility-adjusted model, all \BIQAE{} 95\%
credible intervals cover the true amplitude, whereas the standard
likelihood fails at several steps.  The boundary-amplitude audit shows a
2.2x error reduction, from 0.0086 to 0.0039, after accounting for the
depolarizing floor.

Finally, the calibration protocol is reproducible across backends and
mitigation levels.  The amplitude-0.10 repeat gives the same rounded
error on Kingston without mitigation and on Torino with the mitigation
stack enabled.  That behavior supports the pipeline interpretation
adopted throughout the paper: the calibrated prior is stabilizing
inference itself, while the residual error is dominated by finite
sampling rather than backend-specific artifacts.

%% file: figures/biqae-calibration-compact.tex
\begin{tikzpicture}[
    >=Stealth,
    font=\sffamily,
    stage/.style={
        draw=quantum-blue, fill=quantum-blue!8, thick,
        rounded corners=2pt, minimum width=1.65cm, minimum height=0.58cm,
        font=\scriptsize\sffamily, align=center, inner sep=2pt
    },
    prior/.style={
        draw=classical-red, fill=classical-red!8, thick,
        rounded corners=2pt, minimum width=1.65cm, minimum height=0.58cm,
        font=\scriptsize\sffamily, align=center, inner sep=2pt
    },
    output/.style={
        draw=black!55, fill=black!4, thick,
        rounded corners=2pt, minimum width=1.65cm, minimum height=0.58cm,
        font=\scriptsize\sffamily, align=center, inner sep=2pt
    },
    arr/.style={->, thick, black!70},
    label/.style={font=\tiny\sffamily, text=black!55, align=center},
]

\node[stage] (coarse) at (0, 0) {coarse scan\\depth zero};
\node[output] (regime) at (1.95, 0) {regime\\test};
\node[prior] (nearzero) at (4.05, 0.95) {near-zero\\prior};
\node[prior] (interior) at (4.05, 0) {interior\\Jeffreys};
\node[prior] (nearone) at (4.05, -0.95) {near-one\\prior};
\node[stage] (biqae) at (6.2, 0) {calibrated\\\BIQAE{}};
\node[output] (estimate) at (8.15, 0) {estimate + CI\\for update};

\draw[arr] (coarse) -- node[label, above] {coarse estimate} (regime);
\draw[arr] (regime) -- node[label, above left] {near zero} (nearzero.west);
\draw[arr] (regime) -- (interior.west);
\draw[arr] (regime) -- node[label, below left] {near one} (nearone.west);
\draw[arr] (nearzero.east) -- (biqae.north west);
\draw[arr] (interior.east) -- (biqae.west);
\draw[arr] (nearone.east) -- (biqae.south west);
\draw[arr] (biqae) -- (estimate);

\node[label, text=black!65] at (4.05, -1.55)
  {informative prior avoids wasted deep Grover levels near boundaries};

\end{tikzpicture}

%% file: figures/biqae-boundary-results.tex
\begin{tikzpicture}
\begin{axis}[
    width=7.9cm,
    height=4.9cm,
    symbolic x coords={0.01,0.10,0.50,0.90,0.95},
    xtick=data,
    xlabel={True amplitude},
    ylabel={Absolute estimation error},
    ymode=log,
    ymin=0.001, ymax=1,
    grid=both,
    grid style={black!8},
    major grid style={black!12},
    legend style={
        at={(0.5,-0.24)},
        anchor=north,
        legend columns=2,
        draw=none,
        fill=none,
        font=\scriptsize
    },
    tick label style={font=\scriptsize},
    label style={font=\scriptsize},
]
\addplot[
    color=classical-red,
    mark=square*,
    thick
] coordinates {
    (0.01,0.48)
    (0.10,0.346)
    (0.50,0.007)
    (0.90,0.344)
    (0.95,0.45)
};
\addlegendentry{Uncalibrated baseline}

\addplot[
    color=quantum-blue,
    mark=*,
    thick
] coordinates {
    (0.01,0.002)
    (0.10,0.009)
};
\addlegendentry{Boundary-aware calibration}
\addplot[
    color=quantum-blue,
    mark=*,
    thick,
    forget plot
] coordinates {
    (0.90,0.013)
    (0.95,0.005)
};
\end{axis}
\end{tikzpicture}

%% file: sections/06-nisq-feasibility.tex
\section{Experimental Evaluation}
\label{sec:nisq-feasibility}

Sections~\ref{sec:multistep-grover} and~\ref{sec:biqae-calibration}
establish the core result on Heron~R2; this section adds
supplementary evidence on transferability and operating regime.

\subsection{Hardware Roles and Resource Accounting}
\label{ssec:resource-accounting}

The Heron devices are not interchangeable repetitions.  Kingston
provides the main 8-step trajectory, rare-event envelope, and
mitigation controls; Fez provides the longer 12-step, 20-step, and
32-step horizon extensions; Pittsburgh and Boston are used for Heron~R3
transfer checks; Marrakesh and Torino appear only in supporting
cross-backend or calibration rows.  We keep these roles explicit so
the reader can distinguish a primary result from a transfer or
stress-control run.

\begin{table}[t]
\centering
\caption{Resource accounting for the sequential hardware controls.
The table records the budget shape used by the paper; execution
identifiers are kept in the artifact manifest rather than in the main
text.}
\label{tab:resource-accounting}
\footnotesize
\setlength{\tabcolsep}{2.5pt}
\resizebox{\columnwidth}{!}{%
\begin{tabular}{@{}l c c c@{}}
\toprule
Run & Backend role & Circuits & Shots/step \\
\midrule
8-step No-AA / Grover & Kingston main path & 8 & about 10k \\
8-step matched \FPAA{} & Kingston control & 8 & 10k \\
8-step headline \FPAA{} & Kingston fidelity & 8 & 32\,768 \\
12-step repeat \FPAA{} & Fez horizon & 12 & 8\,192 \\
20-step \FPAA{} & Fez/Kingston transfer & 20 & 8\,192 \\
32-step \FPAA{} & Fez horizon & 32 & 8\,192 \\
\BIQAE{} boundary pairs & Pittsburgh calibration & 2--4 depths & 200+300 / 2\,400 \\
\bottomrule
\end{tabular}
}
\end{table}

Classical reference cost is not the bottleneck in the two-state Tiger
unit cell: the exact Bayes update is closed form and is used as the
posterior oracle.  The relevant comparison is therefore sample cost per
accepted posterior and posterior/action fidelity, not end-to-end
runtime.  Queue-inclusive and queue-exclusive wall-clock timings were
not stored consistently enough across the historical artifacts to
support a runtime-speedup claim.  A future live-hardware rerun should
record submit time, queue entry time, execution start/finish,
calibration snapshot, and IBM Runtime usage fields before making that
comparison.

\subsection{Heron R3 Transfer Validation}
\label{ssec:heron-r3-transfer}

The main campaign remains the Heron~R2 study of
Section~\ref{sec:multistep-grover}; we rerun the unit cell on newer
Heron~R3 processors to check transfer.

\begin{table}[t]
  \centering
  \caption{Supplementary Heron~R3 transfer validation.}
  \label{tab:heron-r3-refresh}
  \footnotesize
  \setlength{\tabcolsep}{2pt}
  \begin{tabular}{@{}>{\raggedright\arraybackslash}p{0.24\columnwidth}
                  >{\raggedright\arraybackslash}p{0.24\columnwidth}
                  c
                  >{\raggedright\arraybackslash}p{0.34\columnwidth}@{}}
    \toprule
    \textbf{Experiment} & \textbf{Backend(s)} & \textbf{Qubits} &
    \textbf{Transfer result} \\
    \midrule
    Same-backend 12-step \FPAA{} trajectory
      & IBM Pittsburgh & 2
      & 12/12 steps pass; max Hellinger 0.0274, mean 0.0077 \\
    Stepwise 12-step continuation
      & Boston steps 0--3; Pittsburgh steps 4--11
      & 2
      & 12/12 steps pass; max Hellinger 0.0148, mean 0.0073 \\
    Same-backend 8-step \FPAA{} trajectory
      & IBM Pittsburgh & 2
      & 8/8 steps pass; max Hellinger 0.0154, mean 0.0076 \\
    4-state optimized six-case sweep
      & IBM Pittsburgh & 3
      & 6/6 pass below Hellinger 0.05; range 0.0129--0.0342 \\
    4-state boundary reruns
      & IBM Pittsburgh & 3
      & near-left prior plus hear-right: Hellinger 0.0438 (4\,096), 0.0342 (8\,192) \\
    \bottomrule
  \end{tabular}
\end{table}

Table~\ref{tab:heron-r3-refresh} is framed as transfer validation, not
as a replacement for the main campaign.  The reading point is that the
loop survives same-backend Pittsburgh repeats, a Boston-to-Pittsburgh
continuation, and a stricter 4-state corridor without turning those
controls into the primary claim.

\subsection{4-State Extension and Boundary Envelope}
\label{ssec:4state-envelope}

The optimised 4-state circuit passes all six tested cases below the
strict Hellinger corridor.  A deeper stress envelope passes three of six
strict-band cases and preserves the MAP state in five of six, marking
the current hardware boundary.

\subsection{Noise Mitigation, Adaptive FPAA, and Hardware Regime}
\label{ssec:diagnostics-regime}

The mitigation and adaptive-\FPAA{} runs are best read as diagnostics,
not as new headline claims.  Layerwise Richardson extrapolation,
bounded zero-noise extrapolation, gate twirling, and adaptive barrier
\FPAA{} all point to the same operating picture: shallow two- and
three-qubit belief oracles can be stabilized, but deeper compiled
circuits quickly become coherence-limited.  Gate twirling helps in the
shallow Tiger and 4-state corridor regime, yet becomes counterproductive
on deeper UCGate chains where the mitigation overhead itself starts to
dominate.  This is why the paper reports a hardware envelope rather than
a blanket mitigation recipe.

Classical tensor-network simulations are also part of the boundary
story.  They make the small state spaces in this study classically
tractable~\cite{tindall_bp_heavyhex_2024,begusic_pauli_propagation_2024,mauron_carleo_dwave_2025},
so the claim is not wall-clock advantage on Tiger.  The relevant target
is the rare-observation sampling bottleneck, a regime where belief
propagation itself can become fragile near critical points of the
underlying model~\cite{midha_bp_tn_limits_2026,aghamohammadi_rare_event_2018}.

\subsection{Scaling Pathway Controls}
\label{ssec:scaling-frontier}

Extended controls beyond the core 8-step to 12-step campaign support the
same interpretation without expanding the primary claim.  The longer
Tiger horizons check that the service does not fail immediately when the
posterior is fed back for many steps.  The 4-state and UCGate pilots ask
a different question: whether state-space size or circuit depth is the
first bottleneck on current Heron processors.  The answer from these
pilots is depth.  When the encoding is shallow enough, the posterior can
remain MAP-stable; when the compiled oracle becomes deep, the run turns
into a noise-frontier marker rather than a validated corridor.

That distinction is the scale-up message.  Larger operational
\POMDP{}s will need richer transition and observation oracles.  The
present data show that state-space size alone is not the limiting
factor, but deeper problem-specific belief encodings remain outside the
validated NISQ corridor and are left to future hardware generations.
Full scaling data and transpile manifests remain in the public artifact
bundle for audit.

%% file: sections/07-conclusion.tex
\section{Implications for Autonomous Systems and Conclusion}
\label{sec:conclusion}

The arXiv revision supports a focused conclusion: \QANTIS{} is best
understood as a calibrated belief-update service inside a classical
autonomous decision stack.  The planner still receives an ordinary
posterior distribution.  The quantum processor is used only inside the
rare-observation evidence update, where amplification can reduce the
sampling burden and boundary-aware \BIQAE{} keeps the estimator from
drifting when the belief becomes very concentrated.

The most important systems check is not the smallest Hellinger number;
it is whether the returned posterior would change the planner's action.
Under the standard Tiger immediate-reward rule, the reported hardware
and exact Bayes posteriors select the same action on the 8-step,
20-step, and 32-step decision checks, with zero measured cumulative
value loss.  The rare-event sweep and boundary calibration explain why
the service can be useful, but they are deliberately framed as operating
envelope evidence rather than a wall-clock speedup or deep-circuit
fidelity claim.

The remaining limitation is also clear.  The current artifacts capture
backend, execution identifiers, circuits, shots, ISA depth, posterior
metrics, raw counts, calibration windows, and transpile manifests, but not reliable
queue-inclusive and queue-exclusive timing.  Future live-hardware reruns
should record those fields before making runtime-cost claims.  The
scale-up evidence points to circuit depth per belief update as the
binding constraint for current Heron devices; richer operational
\POMDP{}s will require shallower problem-specific encodings or future
hardware generations.  Table~\ref{tab:operational-reading} closes the
paper with the operational reading of the evidence: what the hardware
results establish for a systems reader, how the posterior service should
be used, and which measurements remain outside the present claim.
Public code:
\url{https://github.com/neuraparse/qantis}.

\begin{table*}[!t]
\centering
\caption{Operational reading of the \QANTIS{} validation results.  The
table summarizes what each experiment establishes for an autonomous
decision stack while keeping deployment limits explicit.}
\label{tab:operational-reading}
\footnotesize
\setlength{\tabcolsep}{4pt}
\renewcommand{\arraystretch}{1.16}
\begin{tabular}{@{}p{0.18\textwidth}p{0.33\textwidth}p{0.39\textwidth}@{}}
\toprule
Validation component & Evidence in this paper & Operational reading \\
\midrule
Heron belief update
  & IBM Heron runs report backend, circuits, shots, ISA depth, raw counts,
    calibration windows, posterior metrics, and exact-Bayes references
  & The quantum routine can be read as a calibrated rare-evidence posterior
    service; the classical planner interface remains an ordinary posterior
    distribution. \\
Decision evidence
  & Hardware and exact Bayes posteriors, Hellinger distance, MAP agreement,
    and value-loss checks for the 8-, 20-, and 32-step decision tests
  & The systems criterion is action preservation under the stated Tiger reward
    rule, so posterior accuracy is tied to the behavior the controller would
    actually choose. \\
Rare-event regime
  & FPAA posterior error and rare-event lift are measured across concentrated
    evidence settings where naive sampling becomes fragile
  & The useful regime is narrow but important: low-probability observations
    can be amplified without asking the whole autonomous loop to become
    quantum. \\
Boundary calibration
  & Boundary-aware \BIQAE{} and transfer checks keep the estimator stable when
    the posterior mass approaches the edge of the probability simplex
  & Calibration is part of the service, not a cosmetic post-process; it is what
    keeps a concentrated belief from becoming a misleading control signal. \\
Scale-up probes
  & The 4-step UCGate and 12-step Tiger pilots preserve backend, circuit,
    depth, and posterior summaries for the same artifact path
  & These runs localize the next bottleneck as circuit depth per belief
    update; they are controls, not a replacement for the 8-step primary
    hardware result. \\
Runtime boundary
  & The next live-hardware rerun should add submit time, queue entry,
    execution start/finish, calibration snapshot, and runtime usage fields
  & The present paper supports posterior-fidelity and operating-envelope
    claims; wall-clock runtime comparisons should wait for those timing
    fields. \\
\bottomrule
\end{tabular}
\end{table*}